\documentclass[11pt]{article}

\usepackage{acl}

\usepackage{times}
\usepackage{latexsym}

\usepackage[T1]{fontenc}

\usepackage[utf8]{inputenc}

\usepackage{microtype}

\usepackage[scaled=0.8]{beramono}

\usepackage{array}
\usepackage{multirow}
\usepackage{booktabs} 
\usepackage{multicol}

\usepackage{footnote}
\usepackage{fnpct} 

\usepackage[normalem]{ulem} 
\usepackage{graphicx}
\usepackage{subcaption}
\usepackage{tikz,pgf}
\usepackage{tikz-dependency}
\usetikzlibrary{arrows, automata}

\usepackage{amssymb}	
\usepackage{amsmath}
\usepackage{enumitem}

\usepackage{tipa}
\usepackage{adjustbox}

\newcommand{\ensuretext}[1]{#1}
\newcommand{\nertcomment}[4]{\ensuretext{\textcolor{#3}{[\ensuretext{\textcolor{#3}{\ensuremath{^{\textsc{#1}}_{\textsc{#2}}}}} #4]}}}

\newcommand\blfootnote[1]{%
  \begingroup
  \renewcommand\thefootnote{}\footnote{#1}%
  \addtocounter{footnote}{-1}%
  \endgroup
}

\title{Revisiting the Effects of Leakage on Dependency Parsing}

\author{Nathaniel Krasner\textsuperscript{*,1}, Miriam Wanner\textsuperscript{*,2}, Antonios Anastasopoulos\textsuperscript{1}\\
\textsuperscript{1}Department of Computer Science, George Mason University\\
\textsuperscript{2}Department of Computer Science, University of Virginia\\
\texttt{natekrasner@gmail.com, msw2vg@virginia.edu, antonis@gmu.edu}}

\date{}

\begin{document}
\maketitle
\begin{abstract}
Recent work by \citet{sogaard-2020-languages} showed that, treebank size aside, overlap between training and
test graphs (termed \textit{leakage}) explains more of the observed variation in dependency parsing performance than other explanations. \blfootnote{\textsuperscript{*} Equal contribution. Work performed at GMU.}
In this work we revisit this claim, testing it on more models and languages. We find that it only holds for zero-shot cross-lingual settings. We then propose a more fine-grained measure of such leakage which, unlike the original measure, not only explains but also correlates with observed performance variation.\footnote{Code and data are available here: \url{https://github.com/miriamwanner/reu-nlp-project}}
\end{abstract}

\section{Introduction}
\label{sec:intro}

Syntactic parsing has long been one of the core natural language processing (NLP) tasks, and the proliferation of the Universal Dependencies project~\cite[UD;][]{de2021universal,nivre2017universal} has allowed the development and comparison of monolingual and multilingual models under the same syntactic framework.

The performance of the dependency parsers, however, varies wildly across languages, with state-of-the-art performance ranging from labeled attachment scores below 20 (e.g. for Amharic, Erzya, Komi, or Yoruba) to more than 90 (e.g. for Spanish, Polish, Russian, or Greek).
As the UD treebanks follow mostly similar annotation guidelines, comparisons of the parsing performance across languages are now possible, to an extent.\footnote{Different treebank creation protocols followed across languages (whose effects are hard to isolate or measure) can be a significant source of variation. Nevertheless, \textit{some} of the observed variation can be possibly explained by other factors. We direct the reader to footnote 2 of \cite{sogaard-2020-languages}.} 
\begin{figure}
    \centering
    \footnotesize
    \begin{tabular}{cc}
    \multicolumn{2}{l}{Dependency Trees:} \\
\begin{dependency}[row sep=0.5ex,edge unit distance=0.5ex]
  \begin{deptext}[column sep=0.6em]
    She \& saw \& it \\
    PRON \& VERB \& PRON \\
  \end{deptext}
  \deproot[edge unit distance=1ex]{2}{root}
  \depedge{2}{1}{nsubj}
  \depedge{2}{3}{dobj}
\end{dependency}
&
\begin{dependency}[row sep=0.5ex,edge unit distance=0.5ex]
  \begin{deptext}[column sep=0.6em]
    The \& big \& boat \\
    DET \& ADJ \& NOUN \\
  \end{deptext}
  \deproot[edge unit distance=1.2ex]{3}{root}
  \depedge[edge unit distance=1.1ex]{3}{1}{det}
  \depedge{3}{2}{amod}
\end{dependency}
\\

\multicolumn{2}{l}{\citet{sogaard-2020-languages} Unlabeled Directed Graphs:} \\
\begin{tikzpicture}[
            > = stealth, 
            shorten > = 1pt, 
            auto,
            node distance = 1cm, 
            semithick 
        ]
        \tikzstyle{every state}=[
            draw = black,
            thick,
            fill = white,
            minimum size = 4mm
        ]

        \node[state] (s) { };
        \node[state] (v1) [right of=s] {\textsc{ }};
        \node[state] (v2) [below of=v1] {\textsc{ }};
        \node[state] (v3) [below of=s] {\textsc{ }};
        \path[->] (s) edge node {} (v1);
        \path[->] (v1) edge node {} (v2);
        \path[->] (v1) edge node[sloped,above] {} (v3);
\end{tikzpicture}
& 
\begin{tikzpicture}[
            > = stealth, 
            shorten > = 1pt, 
            auto,
            node distance = 1cm, 
            semithick 
        ]
        \tikzstyle{every state}=[
            draw = black,
            thick,
            fill = white,
            minimum size = 4mm
        ]

        \node[state] (s) {};
        \node[state] (v1) [right of=s] {\textsc{}};
        \node[state] (v2) [below of=v1] {\textsc{}};
        \node[state] (v3) [below of=s] {\textsc{}};
        \path[->] (s) edge node {} (v1);
        \path[->] (v1) edge node {} (v2);
        \path[->] (v1) edge node {} (v3);
\end{tikzpicture} \\
\multicolumn{2}{l}{Node- and Edge-Labeled Directed Graphs:} \\
\begin{tikzpicture}[
            > = stealth, 
            shorten > = 1pt, 
            auto,
            node distance = 1.1cm, 
            semithick 
        ]
        \tikzstyle{every state}=[
            draw = black,
            thick,
            fill = white,
            minimum size = 4mm
        ]

        \node[state] (s) {\footnotesize rt};
        \node[state] (v1) [right of=s] {\footnotesize \textsc{V}};
        \node[state] (v2) [below of=v1] {\footnotesize \textsc{Pr}};
        \node[state] (v3) [below of=s] {\footnotesize \textsc{Pr}};
        \path[->] (s) edge node {} (v1);
        \path[->] (v1) edge node {\footnotesize nsubj} (v2);
        \path[->] (v1) edge node[sloped,above] {\footnotesize dobj} (v3);
\end{tikzpicture}
& 
\begin{tikzpicture}[
            > = stealth, 
            shorten > = 1pt, 
            auto,
            node distance = 1.1cm, 
            semithick 
        ]
        \tikzstyle{every state}=[
            draw = black,
            thick,
            fill = white,
            minimum size = 4mm
        ]

        \node[state] (s) {\footnotesize rt};
        \node[state] (v1) [right of=s] {\footnotesize \textsc{N}};
        \node[state] (v2) [below of=v1] {\footnotesize \textsc{dt}};
        \node[state] (v3) [below of=s] {\footnotesize \textsc{A}};
        \path[->] (s) edge node {} (v1);
        \path[->] (v1) edge node {\footnotesize det} (v2);
        \path[->] (v1) edge node[sloped,above] {\footnotesize amod} (v3);
\end{tikzpicture}
\end{tabular}
\caption{Only labeled reductions produce different graphs for these fundamentally different sentences. Under unlabeled leakage, the two trees do leak. When taking labels into account the two trees belong to different isomorphisms and are not considered ``leaky".}
\label{fig:smallexample}
\vspace{-1em}
\end{figure}

In an effort to explain these cross-lingual performance differences, researchers have proposed treebank size~\cite{vania-etal-2019-systematic}, linguistic variation~\cite{nivre-etal-2007-conll}, test data sentence length or average gold
dependency length~\cite{mcdonald-nivre-2011-analyzing}, and domain differences between training and test data~\cite{foster-etal-2011-news}, as potential predictors.  Recently, \citet{sogaard-2020-languages} proposed that the proportion of isomorphic graph structures between the training and testing data (\textit{leakage}) is a stronger predictor of the parsers' performance than any of the previously listed attributes other than training treebank size.

\citet{sogaard-2020-languages} concludes that ``some languages seem easier to parse because their treebanks leak.'' This finding is potentially crucial  for current parser evaluation on the existing treebanks, as well as for future treebank construction. It implies, for instance, that parsers are perhaps not as good as they seem, because they are tested on ``leaky'' test data. Perhaps one should also consider designing treebanks that do not leak between train and test, as such a test set would not have a bias toward more common phenomena.




In this work, we examine this finding more closely. We extend S{\o}gaard's definition to include labeled leakage, and study it over multiple parsers in both monolingual and cross-lingual settings. We show that the finding does not hold up when tested against more modern parsers and more languages. We do identify, though, that leakage indeed predicts parser performance in zero-shot cross-lingual settings, and we dive deeper in this phenomenon with an extensive study focusing on Faroese and other Germanic languages. Last, we propose a modification of the leakage measure that both predicts \textit{and correlates with} parser performance in such settings.



\section{Leakage and How to Measure it}
\label{sec:metrics}

\iftrue
In this section we first define leakage based on graph isomorphisms and reproduce S{\o}gaard's experiments. 
We then show that parsers make \textit{local} decisions that allow them to generalize to unseen graphs, and explore additional measures of leakage, studying whether they help explain parser performance.
Last, we argue that sub-trees are more meaningful units than label-free, tree-level representations. 
\fi

\paragraph{Leakage Definition}

Leakage can be broadly defined as the portion of test trees that have isomorphic counterparts in the train set. 
While dependency trees are labeled, directed graphs with labels both on the nodes and on the edges, \citet{sogaard-2020-languages} performed a reduction by removing labels from both nodes and edges.

Given these reduced graphs, \citet{sogaard-2020-languages} finds the different isomorphisms that are present in the training and the test set, using the VF2 algorithm~\cite{cordella2001improved}.
We note that the isomorphism may or may not rely on node or edge labels. In the experiments below, we perform an ablation between using completely unlabeled directed graphs, node-labeled (but not edge-labeled) directed graphs, and between using the full information of the graphs to compute isomorphisms, namely both node and edge labels.

\begin{table*}[t]
    \centering
    \small
    \begin{tabular}{@{}l|@{ }c@{ }c@{ }c@{ }c@{}}
         \toprule
         \multicolumn{5}{c}{\textbf{Tree-level leakage}} \\
         Leakage & Regression & Expl. & & Spearman's\\
         Attributes & Score & Variance & MAE &  $\rho$ \\
        \midrule
        \multicolumn{5}{l}{\textbf{System: CoNLL'18 (S{\o}gaard)}} \\
        None & 0.162 & 0.143 & 7.257 & -0.194\\
        Edges & 0.091 & -0.121 & 8.592 & -0.181\\
        Nodes+Edges & 0.085 & -0.179 & 8.830 & -0.161\\
        \midrule
        \multicolumn{5}{l}{\textbf{System: UDify}} \\
         None & 0.250 & 0.047 & 13.07 & -0.360\\
        Edges & 0.146 & -0.083 & 14.012 & -0.080\\
        Nodes+Edges & 0.134 & -0.108 & 14.156 & -0.026\\
        \bottomrule
    \end{tabular}
    \hspace{.5cm}
    \begin{tabular}{@{}l|@{ }c@{ }c@{ }c@{ }c@{}}
         \toprule
         \multicolumn{5}{c}{\textbf{Sub-tree-level leakage}} \\
         Leakage & Regression & Expl. & & Spearman's\\
         Attributes & Score & Variance & MAE &  $\rho$ \\
        \midrule
        \multicolumn{5}{l}{\textbf{System: CoNLL'18 (S{\o}gaard)}} \\
        None & 0.054 & -0.137 & 8.248 & -0.238\\
        Edges & 0.083 & -1.202 & 9.444 & 0.390\\
        Nodes+Edges & 0.089 & -0.210 & 8.197 & 0.538\\
        \midrule
        \multicolumn{5}{l}{\textbf{System: UDify}} \\
         None & 0.123 & -0.171 & 14.632 & -0.082\\
        Edges & 0.174 & -0.333 & 14.182 & 0.579\\
        Nodes+Edges & 0.217 & -0.149 & 13.715 & 0.654\\
        \bottomrule
    \end{tabular}
    \caption{Tree-level leakage (left) does not correlate with and does not always explain parser performance. Labeled sub-tree level leakage (right) however is positively correlated with parser performance.}
    \label{tab:results}
    \vspace{-1em}
\end{table*}

\paragraph{Reproducing~\cite{sogaard-2020-languages}}

Examples of the reductions needed for computing leakage for two sentences are shown in Figure~\ref{fig:smallexample}.
%
Now, assume that the first sentence is in the training set and the second is part of the test set. Measuring leakage without labels implies that the first dependency tree is somehow informative for producing the tree for the second sentence, which we believe is counter-intuitive. Hence, our first hypothesis is that a more informed leakage calculation is going to explain more of the performance variance.

We reproduce the experiments of \citet{sogaard-2020-languages} comparing the three different reductions (denoted as "none" for unlabeled graphs, "edges" and "nodes+edges" for respectively labeled graphs).
The experiment consists of correlating the factors $\phi$ assumed to influence syntactic dependency parser performance with the performance of the parser under study.
We train a simple linear regression model\footnote{Exactly as \citet{sogaard-2020-languages} does, just on different data/settings.} with treebank
size and $\phi$ as input and parser performance as output. $\phi$ will correspond to our measure of treebank leakage. Mathematically, we have $\alpha t_s + \beta\phi + \gamma$ with $t_s$ treebank size and $\alpha,\beta,\gamma$ learned parameters. 
Following S{\o}gaard, we will focus on explained variance and mean absolute error (MAE) from five-fold cross-validation to avoid overfitting. Unlike S{\o}gaard, we will additionally report Spearman's $\rho$ correlation coefficients\footnote{We do not expect the correlation, if any, to be linear, hence we prefer Spearman's measure to Pearson's.} between factor and performance, which will reveal whether indeed leakage leads to better parser performance.\footnote{Note that the explained variance is basically the correlation squared. As such, it cannot reveal whether the correlation is positive or negative. Negative explained variance means that the model is a poor fit for the data (worse than just predicting the average).}

The results on the same data as~\citet{sogaard-2020-languages} (using the best reported parser performance from the CoNLL 2018 shared task) are presented in the top three rows of Table~\ref{tab:results}.
We find that unlabeled graph leakage produces positive explained variance, in line with previous work. However, we have to reject our hypothesis, as a more informed leakage measurement fails to meaningfully explain the output variance, producing negative scores. In fact, the more information we use when computing the graph isomorphisms, the less the model can explain output variance!\footnote{\citet{sogaard-2020-languages} gives this possible explanation: ``The result is perhaps not too surprising, since graph isomorphisms correlate with syntactic constructions, which in turn correlate with the occurrence of linguistic markers and tail linguistic phenomena."}

To further solidify this finding, we repeat the above experiment, this time using UDify, the state-of-the-art multilingual parser of~\citet{kondratyuk-straka-2019-udify}.\footnote{The model is trained jointly on all UD treebanks (that have a training set), and hence in this experiment we compute leakage multilingually (i.e. we compute leakage between the complete training set and the test set of each treebank).} 
The result is shown in the bottom three rows of Table~\ref{tab:results} (left Table), and they present more negative evidence for our hypothesis: there is minimal explained variance in the unlabeled version, and still negative explained variance in the labeled leakage versions.

Hence, we have to --for now-- reject our hypothesis: using labeled graph isomorphisms to compute leakage does not explain more downstream parser performance variations, at least when using tree-level leakage measures; we revisit this hypothesis for sub-tree leakage below. 
Concurrently, we need to highlight the fact that for \textit{all} cases we focused on this experiment, there was a negative (inverse) correlation between leakage and parser performance. 

While \citet{sogaard-2020-languages} was correct (for the languages/parsers they studied) to state that there is a correlation between leakage and parser performance, we believe they reached an incorrect conclusion. The metric they used (explained variance) does not reveal the direction of the correlation, just that there is a correlation. Because of this they came to the wrong conclusion that there was a positive correlation between leakage and parser performance. Our results (Table~\ref{tab:results}, left table) instead imply that as leakage increases, parser performance worsens! 
Clearly, something is wrong and we need to re-examine S{\o}gaard's reasoning.


\paragraph{Sub-Trees are More Meaningful Units}

\begin{table}[t]
    \centering
    \small
    \begin{tabular}{@{}l|@{ }c@{ }|cc@{}}
    \toprule
         & & \multicolumn{2}{c}{Produced by model trained without}  \\
        Construction & Actual & \texttt{nsubj} mods & \texttt{obj} mods \\
    \midrule
        \texttt{nsubj} mods  & 3166 & 1698 & 3167 \\
        \texttt{obj} mods & 3910 & 4505 & 1940 \\
    \bottomrule
    \end{tabular}
    \caption{Number of adjectival modifiers produced by counterfactual models. 
    The parsers can produce constructions not seen during training.}
    \label{tab:localparse}
    \vspace{-1em}
\end{table}

We turn our attention to the parsers whose performance we are trying to explain. 

The three parsers that S{\o}gaard uses and UDify are graph-based ones. 
This means that they do not necessarily score or produce whole trees. Graph-based parsers score pairs of words, and from these scores a minimum spanning tree is selected to produce the final dependency parse. As such, we argue that whole trees as a measure of leakage are not appropriate for graph-based parsers.

To drive this point forward, we perform synthetic experiments removing adjectival modifiers from nominal subject or object.
In particular, we created training data in which the data either did not contain adjectival modifiers on subject/object nouns. We then tested the models on gold unmodified test data which contained such modifiers. 
By removing adjectival modifiers only from the subjects (similarly for objects) in the training data, we ensure two things: that test instances with adjectival modifiers on subjects are not leaky; as such, if whole-tree leakage is a proper indication for parser performance, then the parser should perform poorly in producing such constructions. 

Table~\ref{tab:localparse} shows the results of our experiment over the German HDT treebank.\footnote{Also see results on English in Appendix~\ref{app:english}.} We found that the parsers trained on our counterfactual data, which have zero leakage for these test instances, still produce the local constructions that they have never observed during training. The parsers trained without training subject modifiers produced about half of the expected subject modifiers (similarly for object modifier experiments). Nevertheless, they were still able to generalize based on other similar constructions seen in training, correctly parsing a non-zero amount of unseen-in-training constructions. 

This observation is not unexpected. In the experiment above, even removing all adjectival modifiers from nouns that are subjects (hence a subtree --and consequently a whole tree containing it-- has never been observed), the parser still observes adjective-noun modifying pairs elsewhere in the sentences and is able to generalize, producing a tree that has never been observed at training.

The fact that the parsers make \textit{local} output decisions, along with the proven corollary that they can easily produce unseen trees, guides us to search for a leakage measure focusing on sub-trees.

\paragraph{Sub-Tree Based Leakage}

\begin{figure}
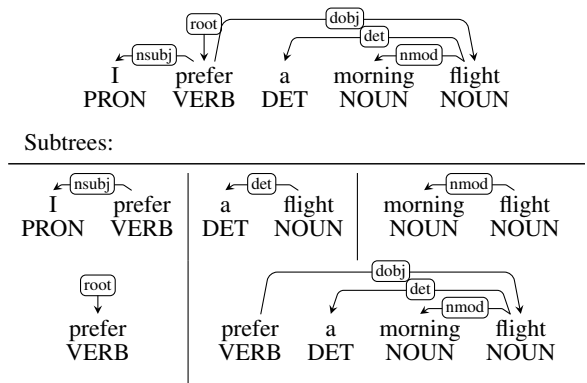

    \centering
    \footnotesize
    \begin{tabular}{@{}c@{}|@{}c@{}|@{}c@{}}
\multicolumn{3}{c}{
\begin{dependency}[row sep=0.5ex,edge unit distance=0.5ex]
  \begin{deptext}[column sep=0.6em]
    I \& prefer \& a \& morning \& flight \\
    PRON \& VERB \& DET \& NOUN \& NOUN \\
  \end{deptext}
  \deproot[edge unit distance=1.2ex]{2}{root}
  \depedge{2}{1}{nsubj}
  \depedge[edge unit distance=1.3ex]{2}{5}{dobj}
  \depedge[edge unit distance=1.1ex]{5}{3}{det}
  \depedge[edge unit distance=0.5ex]{5}{4}{nmod}
\end{dependency}} \\
\multicolumn{3}{l}{Subtrees:} \\
\midrule
\begin{dependency}[row sep=0.5ex,edge unit distance=0.5ex]
  \begin{deptext}[column sep=0.6em]
    I \& prefer \\
    PRON \& VERB \\
  \end{deptext}
  \depedge{2}{1}{nsubj}
\end{dependency}
&
\begin{dependency}[row sep=0.5ex,edge unit distance=0.5ex]
  \begin{deptext}[column sep=0.6em]
    a \& flight \\
    DET \& NOUN \\
  \end{deptext}
  \depedge{2}{1}{det}
\end{dependency}
&
\begin{dependency}[row sep=0.5ex,edge unit distance=0.5ex]
  \begin{deptext}[column sep=0.6em]
    morning \& flight \\
    NOUN \& NOUN \\
  \end{deptext}
  \depedge{2}{1}{nmod}
\end{dependency}
\\
\begin{dependency}[row sep=0.5ex,edge unit distance=0.5ex]
  \begin{deptext}[column sep=0.6em]
    prefer \\
    VERB \\
  \end{deptext}
  \deproot[edge unit distance=1ex]{1}{root}
\end{dependency}
&
\multicolumn{2}{c}{
\begin{dependency}[row sep=0.5ex,edge unit distance=0.5ex]
  \begin{deptext}[column sep=0.6em]
    prefer \& a \& morning\& flight \\
    VERB \& DET \& NOUN \& NOUN \\
  \end{deptext}
  \depedge[edge unit distance=1.3ex]{1}{4}{dobj}
  \depedge[edge unit distance=1.1ex]{4}{2}{det}
  \depedge[edge unit distance=0.5ex]{4}{3}{nmod}
\end{dependency}}\\
\bottomrule
\end{tabular}
\vspace{-1em}
\caption{Decomposition of a tree (top) into a set of sub-trees (bottom).}
\label{fig:subexample}
\vspace{-1em}
\end{figure}

We define a leakage measure where a dependency tree is first decomposed into a set of subtrees, and then each subtree reduces into the graphs defined above 
to compute isomorphisms. 
These subtrees are created for each node (word), connecting it to its parent and to all its children. See example in Figure~\ref{fig:subexample}.

We repeat our experiments, 
this time using our proposed leakage measure, and present the results in the right-hand side of Table~\ref{tab:results}. As before, for S{\o}gaard's and for the UDify combinations of models/languages the explained variance is negative. However, now more information  (edge/node labels) leads to higher Spearman's $\rho$ coefficients, implying that indeed the more test subtrees we have observed in training, the better the parser performance.

In the unlabeled setting, 
every sub-tree created by the parser was found in the training data, which was true of most gold files as well. We interpret this observation to mean that \textit{unlabeled} sub-trees are not meaningful units, a point further reinforced by the negative explained variance and correlations. 

At the same time, our measure still fails to explain any of the observed performance variance. Thus, we have to reach a conclusion opposite of~\citet{sogaard-2020-languages}, that \textbf{in a monolingual setting} the performance of modern graph-based parsers is not particularly explained by train-test leakage, however we compute that leakage.




\section{Leakage Explains 0-Shot Performance}
\label{sec:zeroshot}
Modern dependency parsing models trained on many languages perform well on languages unseen (zero-shot setting) during training~\cite[\textit{inter alia}]{muller-etal-2021-unseen,glavas-vulic-2021-climbing}. 


We focus again on UDify, 
since it performs well in zero-shot settings. 
This is generally attributed to two factors: the presence of related languages in the training set, and the multilingual capabilities of the underlying representation model (here, a multilingual BERT~\cite{devlin2019bert} model).

\begin{table}[t]
    \centering
    \small
    \begin{tabular}{@{}l|@{ }c@{ }c@{ }c@{ }c@{}}
         \toprule
         System: UDify  & Regression & Expl. & & Spearman's\\
         (Zero-Shot) & Score & Variance & MAE &  $\rho$ \\
        \midrule
        \multicolumn{5}{l}{\textbf{Whole-Tree Leakage}} \\
        None & 0.385 & 0.263 & 17.539 & -0.581 \\
        Edges & 0.221 & 0.124 & 21.460 & -0.493 \\
        Nodes+Edges & 0.221 & 0.106 & 21.506 & -0.546 \\
        \midrule
        \multicolumn{5}{l}{\textbf{Sub-Tree Leakage}} \\
        Edges & 0.271 & 0.210 & 18.245 & 0.609\\
        Nodes+Edges & 0.246 & 0.215 & 18.038 & 0.578\\
        \bottomrule
    \end{tabular}
    \caption{Leakage explains zero-shot parser performance. Sub-tree leakage also correlates with it.}
    \label{tab:zeroshot}
    \vspace{-1em}
\end{table}

Table~\ref{tab:zeroshot} reports results with S{\o}gaard's (whole-tree) and our sub-tree leakage measures under all three settings, focusing only on 35 zero-shot test languages.\footnote{These 35 languages are treebanks with size $0$ in \citet{kondratyuk-straka-2019-udify}.} Now leakage\footnote{In this multilingual setting, leakage is computed against a training set comprised of 75 concatenated treebanks.} can indeed explain downstream parser performance. 
Our proposed measure explains as much variance as the original whole-tree measure \textit{and also} correlates with performance. 

\begin{figure*}[t]
    \centering
    \includegraphics[width=.9\textwidth]{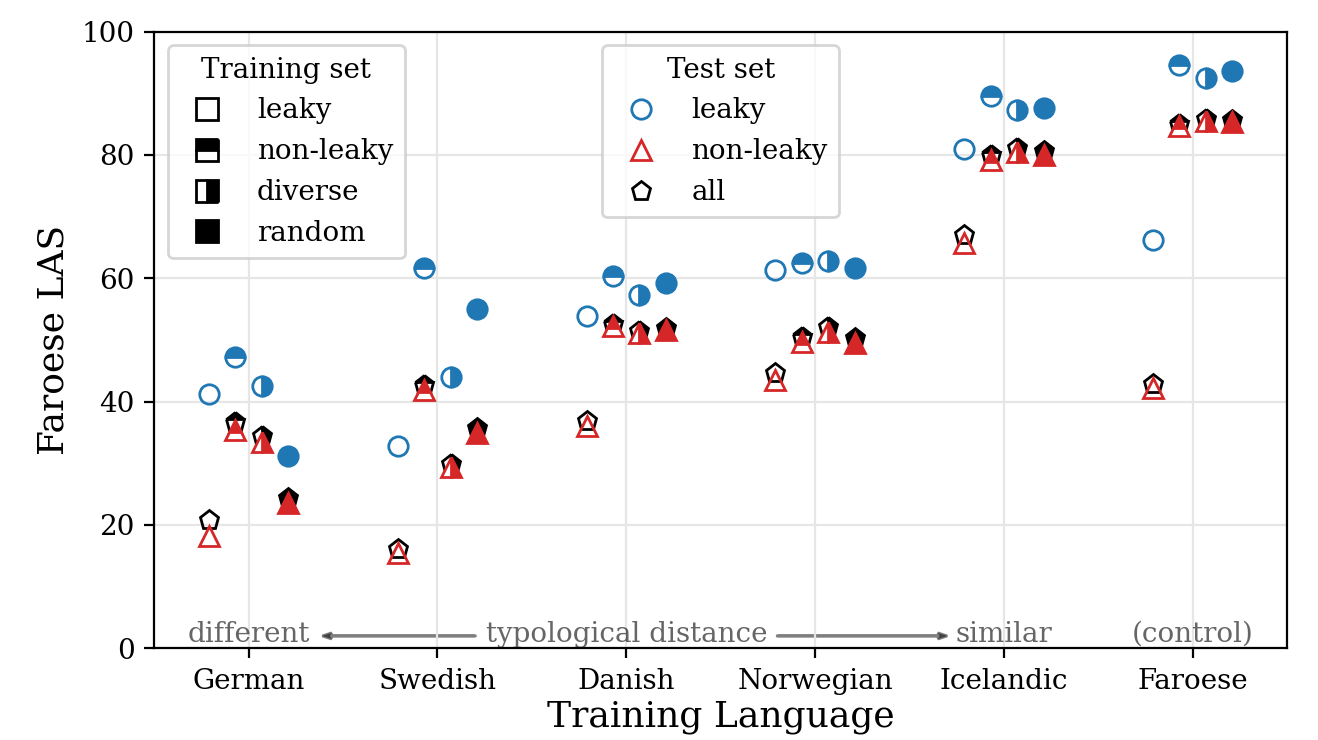}
    \caption{Zero-shot results on Faroese. Training on non-leaky and diverse data is best. The leaky portion of the test set is far easier than the rest.}
    \label{fig:faroese}
    \vspace{-1em}
\end{figure*}

\paragraph{Analysis on Faroese}

Looking deeper into the zero-shot setting, we perform an experiment on a simplified bilingual zero-shot setting. We train parsers in five Germanic languages (German, Swedish, Danish, Norwegian, Icelandic) and test on Faroese in a zero-shot fashion. 
For each language, we train a model on: 

\iftrue
\begin{itemize}[noitemsep, nolistsep]
    \item[(a)] a `leaky' sample of the portion of training treebank, so that all training data overlapped with (some) test data,
    \item[(b)] a `non-leaky' sample of trees such that there was no train-test overlap,
    \item[(c)] a control random sample from the training treebank, and 
    \item[(d)] a `diverse' training sample including a single tree from each isomorphism equivalence class.
\end{itemize}
\fi
All of the above models are size-controlled for each language, so that the  training data sizes are exactly the same. Leakage here is measured with unlabeled full-tree leakage, for simplicity.
We similarly split the test set, for each language, into `leaky' and `non-leaky' subsets (also reporting numbers for the whole test set). 

For example, take German training and Faroese test sets. First all German instances leaking into Faroese are added to the ``German leaky'' train set and the corresponding leaked Faroese sentences are put into the ``Faroese-leaky" test set. Then the remaining sentences from the German training set are added to the ``German nonleaky'' train set and the remaining sentences from the Faroese test set are the ``Faroese nonleaky'' test set. Last, we take a random sample of same size across all settings, so that training data size is not a confounding factor for our analysis. 



See Figure~\ref{fig:faroese} and extensive results in Appendix~\ref{app:faroese}.
For all languages, models trained on leaky data perform worse than models trained on the same amount of non-leaky or random data. For most transfer languages, in fact, training solely on non-leaky data performs better than training on other subsets! In addition, the leaky part of the testing data is clearly \textit{easier} to parse in general, while the non-leaky part is more challenging. 

The models trained on perfectly `diverse' treebanks generally perform just as good as those trained on non-leaky or randomly sampled data and often better on the non-leaky test set, which means they generalize better. This indicates a way to reach better cross-lingual performance without the need for large training data, as long as the training set is diverse enough.

The large performance difference between models trained on leaky and non-leaky trees reveals that something is different about the parts of the treebanks that leak. 
We measured the diversity of the leaky, non-leaky, and randomly selected trees, defined as the number of unique trees divided by the total number of trees. We found that leaky treebanks were far less diverse and therefore contain fewer unique structures than non-leaky or randomly sampled counterparts. Across all treebanks, leaky instances are also generally shorter (8.4 vs 21.6 avg length), shallower (2.2 vs 4.8 average tree depth), and with shorter avg dependency length (2 vs 3.3). 

We argue that the reasoning should be reverse: short ``easy'' examples are more likely to leak; it is not leakage that makes them easy!


\section{Conclusion}
\label{sec:conclusion}

We re-evaluate claims that training-test leakage can explain parser performance, define a subtree-based leakage measure that better explains performance, and show that this claim only holds for zero-shot transfer settings. 

\section*{Acknowledgements}
The authors express their gratitude to Djamé Seddah for very helpful initial discussions, and to the reviewers for their constructive feedback. The first two authors were supported in part by the NSF IIS-1757064 Undergraduate Research in Educational Data Mining grant. Antonios Anastasopoulos is also generously supported by NSF grants BCS-2109578 and IIS-2125466.

\bibliographystyle{acl_natbib}
\bibliography{anthology,emnlp2021}

\begin{thebibliography}{12}
\expandafter\ifx\csname natexlab\endcsname\relax\def\natexlab#1{#1}\fi

\bibitem[{Cordella et~al.(2001)Cordella, Foggia, Sansone, and
  Vento}]{cordella2001improved}
Luigi~Pietro Cordella, Pasquale Foggia, Carlo Sansone, and Mario Vento. 2001.
\newblock An improved algorithm for matching large graphs.

\bibitem[{de~Marneffe et~al.(2021)de~Marneffe, Manning, Nivre, and
  Zeman}]{de2021universal}
Marie-Catherine de~Marneffe, Christopher~D Manning, Joakim Nivre, and Daniel
  Zeman. 2021.
\newblock Universal dependencies.
\newblock \emph{Computational linguistics}, 47(2):255--308.

\bibitem[{Devlin et~al.(2019)Devlin, Chang, Lee, and
  Toutanova}]{devlin2019bert}
Jacob Devlin, Ming-Wei Chang, Kenton Lee, and Kristina Toutanova. 2019.
\newblock {BERT}: Pre-training of deep bidirectional transformers for language
  understanding.
\newblock In \emph{Proceedings of the 2019 Conference of the North American
  Chapter of the Association for Computational Linguistics: Human Language
  Technologies, Volume 1 (Long and Short Papers)}, pages 4171--4186.

\bibitem[{Foster et~al.(2011)Foster, {\c{C}}etino{\u{g}}lu, Wagner, Le~Roux,
  Nivre, Hogan, and van Genabith}]{foster-etal-2011-news}
Jennifer Foster, {\"O}zlem {\c{C}}etino{\u{g}}lu, Joachim Wagner, Joseph
  Le~Roux, Joakim Nivre, Deirdre Hogan, and Josef van Genabith. 2011.
\newblock \href {https://www.aclweb.org/anthology/I11-1100} {From news to
  comment: Resources and benchmarks for parsing the language of web 2.0}.
\newblock In \emph{Proceedings of 5th International Joint Conference on Natural
  Language Processing}, pages 893--901, Chiang Mai, Thailand. Asian Federation
  of Natural Language Processing.

\bibitem[{Glava{\v{s}} and Vuli{\'c}(2021)}]{glavas-vulic-2021-climbing}
Goran Glava{\v{s}} and Ivan Vuli{\'c}. 2021.
\newblock \href {https://doi.org/10.18653/v1/2021.findings-acl.431} {Climbing
  the tower of treebanks: Improving low-resource dependency parsing via
  hierarchical source selection}.
\newblock In \emph{Findings of the Association for Computational Linguistics:
  ACL-IJCNLP 2021}, pages 4878--4888, Online. Association for Computational
  Linguistics.

\bibitem[{Kondratyuk and Straka(2019)}]{kondratyuk-straka-2019-udify}
Dan Kondratyuk and Milan Straka. 2019.
\newblock 75 languages, 1 model: Parsing universal dependencies universally.
\newblock In \emph{Proceedings of the 2019 Conference on Empirical Methods in
  Natural Language Processing and the 9th International Joint Conference on
  Natural Language Processing (EMNLP-IJCNLP)}, pages 2779--2795.

\bibitem[{McDonald and Nivre(2011)}]{mcdonald-nivre-2011-analyzing}
Ryan McDonald and Joakim Nivre. 2011.
\newblock \href {https://doi.org/10.1162/coli_a_00039} {Analyzing and
  integrating dependency parsers}.
\newblock \emph{Computational Linguistics}, 37(1):197--230.

\bibitem[{Muller et~al.(2021)Muller, Anastasopoulos, Sagot, and
  Seddah}]{muller-etal-2021-unseen}
Benjamin Muller, Antonios Anastasopoulos, Beno{\^\i}t Sagot, and Djam{\'e}
  Seddah. 2021.
\newblock \href {https://doi.org/10.18653/v1/2021.naacl-main.38} {When being
  unseen from m{BERT} is just the beginning: Handling new languages with
  multilingual language models}.
\newblock In \emph{Proceedings of the 2021 Conference of the North American
  Chapter of the Association for Computational Linguistics: Human Language
  Technologies}, pages 448--462, Online. Association for Computational
  Linguistics.

\bibitem[{Nivre et~al.(2017)Nivre, Agi{\'c}, Ahrenberg, Antonsen, Aranzabe,
  Asahara, Ateyah, Attia, Atutxa, Augustinus et~al.}]{nivre2017universal}
Joakim Nivre, {\v{Z}}eljko Agi{\'c}, Lars Ahrenberg, Lene Antonsen, Maria~Jesus
  Aranzabe, Masayuki Asahara, Luma Ateyah, Mohammed Attia, Aitziber Atutxa,
  Liesbeth Augustinus, et~al. 2017.
\newblock Universal dependencies 2.1.

\bibitem[{Nivre et~al.(2007)Nivre, Hall, K{\"u}bler, McDonald, Nilsson, Riedel,
  and Yuret}]{nivre-etal-2007-conll}
Joakim Nivre, Johan Hall, Sandra K{\"u}bler, Ryan McDonald, Jens Nilsson,
  Sebastian Riedel, and Deniz Yuret. 2007.
\newblock \href {https://www.aclweb.org/anthology/D07-1096} {The {C}o{NLL} 2007
  shared task on dependency parsing}.
\newblock In \emph{Proceedings of the 2007 Joint Conference on Empirical
  Methods in Natural Language Processing and Computational Natural Language
  Learning ({EMNLP}-{C}o{NLL})}, pages 915--932, Prague, Czech Republic.
  Association for Computational Linguistics.

\bibitem[{S{\o}gaard(2020)}]{sogaard-2020-languages}
Anders S{\o}gaard. 2020.
\newblock \href {https://doi.org/10.18653/v1/2020.emnlp-main.220} {Some
  languages seem easier to parse because their treebanks leak}.
\newblock In \emph{Proceedings of the 2020 Conference on Empirical Methods in
  Natural Language Processing (EMNLP)}, pages 2765--2770, Online. Association
  for Computational Linguistics.

\bibitem[{Vania et~al.(2019)Vania, Kementchedjhieva, S{\o}gaard, and
  Lopez}]{vania-etal-2019-systematic}
Clara Vania, Yova Kementchedjhieva, Anders S{\o}gaard, and Adam Lopez. 2019.
\newblock \href {https://doi.org/10.18653/v1/D19-1102} {A systematic comparison
  of methods for low-resource dependency parsing on genuinely low-resource
  languages}.
\newblock In \emph{Proceedings of the 2019 Conference on Empirical Methods in
  Natural Language Processing and the 9th International Joint Conference on
  Natural Language Processing (EMNLP-IJCNLP)}, pages 1105--1116, Hong Kong,
  China. Association for Computational Linguistics.

\end{thebibliography}

\clearpage
\newpage
\appendix

\section{Graph Reduction Examples}
\label{app:example}
Shown in Table~\ref{fig:example}.
\begin{figure}[h]
    \centering
    \footnotesize
    \begin{tabular}{cc}
    \multicolumn{2}{l}{Dependency Trees:} \\
\begin{dependency}[row sep=0.5ex]
  \begin{deptext}[column sep=0.6em]
    She \& saw \& it \\
    PRON \& VERB \& PRON \\
  \end{deptext}
  \deproot[edge unit distance=2ex]{2}{root}
  \depedge{2}{1}{nsubj}
  \depedge{2}{3}{dobj}
\end{dependency}
&
\begin{dependency}[row sep=0.5ex]
  \begin{deptext}[column sep=0.6em]
    The \& big \& boat \\
    DET \& ADJ \& NOUN \\
  \end{deptext}
  \deproot[edge unit distance=2ex]{3}{root}
  \depedge{3}{1}{det}
  \depedge{3}{2}{amod}
\end{dependency}
\\
\\

\multicolumn{2}{l}{\citet{sogaard-2020-languages} Unlabeled Directed Graphs:} \\
\begin{tikzpicture}[
            > = stealth, 
            shorten > = 1pt, 
            auto,
            node distance = 1cm, 
            semithick 
        ]
        \tikzstyle{every state}=[
            draw = black,
            thick,
            fill = white,
            minimum size = 4mm
        ]

        \node[state] (s) { };
        \node[state] (v1) [right of=s] {\textsc{ }};
        \node[state] (v2) [below of=v1] {\textsc{ }};
        \node[state] (v3) [below of=s] {\textsc{ }};
        \path[->] (s) edge node {} (v1);
        \path[->] (v1) edge node {} (v2);
        \path[->] (v1) edge node[sloped,above] {} (v3);
\end{tikzpicture}
& 
\begin{tikzpicture}[
            > = stealth, 
            shorten > = 1pt, 
            auto,
            node distance = 1cm, 
            semithick 
        ]
        \tikzstyle{every state}=[
            draw = black,
            thick,
            fill = white,
            minimum size = 4mm
        ]

        \node[state] (s) {};
        \node[state] (v1) [right of=s] {\textsc{}};
        \node[state] (v2) [below of=v1] {\textsc{}};
        \node[state] (v3) [below of=s] {\textsc{}};
        \path[->] (s) edge node {} (v1);
        \path[->] (v1) edge node {} (v2);
        \path[->] (v1) edge node {} (v3);
\end{tikzpicture} \\
\multicolumn{2}{l}{Node-Labeled Directed Graphs:} \\
\begin{tikzpicture}[
            > = stealth, 
            shorten > = 1pt, 
            auto,
            node distance = 1cm, 
            semithick 
        ]
        \tikzstyle{every state}=[
            draw = black,
            thick,
            fill = white,
            minimum size = 4mm
        ]

        \node[state] (s) {\footnotesize rt};
        \node[state] (v1) [right of=s] {\footnotesize \textsc{V}};
        \node[state] (v2) [below of=v1] {\footnotesize \textsc{Pr}};
        \node[state] (v3) [below of=s] {\footnotesize \textsc{Pr}};
        \path[->] (s) edge node {} (v1);
        \path[->] (v1) edge node {} (v2);
        \path[->] (v1) edge node {} (v3);
\end{tikzpicture}
& 
\begin{tikzpicture}[
            > = stealth, 
            shorten > = 1pt, 
            auto,
            node distance = 1cm, 
            semithick 
        ]
        \tikzstyle{every state}=[
            draw = black,
            thick,
            fill = white,
            minimum size = 4mm
        ]

        \node[state] (s) {\footnotesize rt};
        \node[state] (v1) [right of=s] {\footnotesize \textsc{N}};
        \node[state] (v2) [below of=v1] {\footnotesize \textsc{dt}};
        \node[state] (v3) [below of=s] {\footnotesize \textsc{A}};
        \path[->] (s) edge node {} (v1);
        \path[->] (v1) edge node {} (v2);
        \path[->] (v1) edge node {} (v3);
\end{tikzpicture} \\
\multicolumn{2}{l}{Node- and Edge-Labeled Directed Graphs:} \\
\begin{tikzpicture}[
            > = stealth, 
            shorten > = 1pt, 
            auto,
            node distance = 1.1cm, 
            semithick 
        ]
        \tikzstyle{every state}=[
            draw = black,
            thick,
            fill = white,
            minimum size = 4mm
        ]

        \node[state] (s) {\footnotesize rt};
        \node[state] (v1) [right of=s] {\footnotesize \textsc{V}};
        \node[state] (v2) [below of=v1] {\footnotesize \textsc{Pr}};
        \node[state] (v3) [below of=s] {\footnotesize \textsc{Pr}};
        \path[->] (s) edge node {} (v1);
        \path[->] (v1) edge node {\footnotesize nsubj} (v2);
        \path[->] (v1) edge node[sloped,above] {\footnotesize dobj} (v3);
\end{tikzpicture}
& 
\begin{tikzpicture}[
            > = stealth, 
            shorten > = 1pt, 
            auto,
            node distance = 1.1cm, 
            semithick 
        ]
        \tikzstyle{every state}=[
            draw = black,
            thick,
            fill = white,
            minimum size = 4mm
        ]

        \node[state] (s) {\footnotesize rt};
        \node[state] (v1) [right of=s] {\footnotesize \textsc{N}};
        \node[state] (v2) [below of=v1] {\footnotesize \textsc{dt}};
        \node[state] (v3) [below of=s] {\footnotesize \textsc{A}};
        \path[->] (s) edge node {} (v1);
        \path[->] (v1) edge node {\footnotesize det} (v2);
        \path[->] (v1) edge node[sloped,above] {\footnotesize amod} (v3);
\end{tikzpicture}
\end{tabular}
\caption{Only labeled reductions produce different graphs for these fundamentally different sentences.}
\label{fig:example}
\end{figure}

\section{English Counterfactual Experiments}
\label{app:english}
Shown in Table~\ref{tab:localboth}.

\begin{table}[h]
    \centering
    \small
    \begin{tabular}{@{}l|@{ }c@{ }|cc@{}}
    \toprule
         & & \multicolumn{2}{c}{Produced by model trained without}  \\
        Construction & Actual & \texttt{nsubj} mods & \texttt{obj} mods \\
    \midrule
    \multicolumn{4}{l}{\textbf{German HDT treebank:}}\\
        \texttt{nsubj} mods  & 3166 & 1698 & 3167 \\
        \texttt{obj} mods & 3910 & 4505 & 1940 \\
    \midrule
    \multicolumn{4}{l}{\textbf{English EWT treebank:}}\\
        \texttt{nsubj} mods  & 132 & 122 & 130 \\
        \texttt{obj} mods & 254 & 258 & 237  \\
    \bottomrule
    \end{tabular}
    \vspace{-1em}
    \caption{Number of adjectival modifiers produced by counterfactual models compared to the actual number in the gold file. 
    The parsers can produce constructions not seen during training.}
    \label{tab:localboth}
\end{table}

\section{Complete Faroese Results}
\label{app:faroese}
Shown in Tables~\ref{tab:faroese}.
\begin{table*}[t]

\begin{tabular}{cc|ccccc}

\toprule
\textbf{Train}              & \textbf{Test}             & \textbf{UAS}   & \textbf{LAS}   & \textbf{CLAS}  & \textbf{MLAS}  & \textbf{BLEX}  \\
\midrule
Faroese-leaky      & Faroese-leaky    & 78.48 & 66.24 & 55.6  & 51.26 & 55.6  \\
                   & Faroese-nonleaky & 53.22 & 42.19 & 28.39 & 22.6  & 28.39 \\
                   & Faroese-all      & 53.91 & 42.85 & 29.19 & 23.43 & 29.19 \\
\midrule
Faroese-nonleaky   & Faroese-leaky    & 97.89 & 94.51 & 90.65 & 90.65 & 90.65 \\
                   & Faroese-nonleaky & 88.15 & 84.75 & 78.66 & 75.98 & 78.66 \\
                   & Faroese-all      & 88.42 & 85.02 & 79.01 & 76.41 & 79.01 \\
\midrule
Faroese-all        & Faroese-leaky    & 97.47 & 93.67 & 90.32 & 88.17 & 90.32 \\
                   & Faroese-nonleaky & 89    & 85.36 & 79.02 & 76.56 & 79.02 \\
                   & Faroese-all      & 89.23 & 85.59 & 79.35 & 76.9  & 79.35 \\
\midrule
Faroese-diverse    & Faroese-leaky    & 96.62 & 92.41 & 89.61 & 85.3  & 89.61 \\
                   & Faroese-nonleaky & 89.27 & 85.56 & 79.66 & 77.2  & 79.66 \\
                   & Faroese-all      & 89.47 & 85.75 & 79.95 & 77.43 & 79.95 \\
\midrule
German-leaky       & Faroese-leaky    & 57.49 & 41.23 & 43    & 34.89 & 43    \\
                   & Faroese-nonleaky & 32.53 & 18.24 & 16.24 & 12.57 & 16.24 \\
                   & Faroese-all      & 35.25 & 20.74 & 19.27 & 15.1  & 19.27 \\
\midrule
German-nonleaky    & Faroese-leaky    & 64.08 & 47.18 & 51.43 & 40.86 & 51.43 \\
                   & Faroese-nonleaky & 52.07 & 35.43 & 35.14 & 27.09 & 35.14 \\
                   & Faroese-all      & 53.38 & 36.71 & 36.98 & 28.65 & 36.98 \\
\midrule
German-all         & Faroese-leaky    & 57.39 & 31.24 & 36.84 & 25.82 & 36.84 \\
                   & Faroese-nonleaky & 45.81 & 23.59 & 25.03 & 19.81 & 25.03 \\
                   & Faroese-all      & 47.07 & 24.42 & 26.36 & 20.49 & 26.36 \\
\midrule
German-diverse     & Faroese-leaky    & 58.02 & 42.51 & 48.94 & 29.73 & 48.94 \\
                   & Faroese-nonleaky & 48.95 & 33.4  & 34.26 & 23.6  & 34.26 \\
                   & Faroese-all      & 49.94 & 34.39 & 35.9  & 24.29 & 35.9  \\
\midrule
Afrikaans-leaky    & Faroese-leaky    & 46.51 & 20.47 & 24.91 & 11.72 & 24.91 \\
                   & Faroese-nonleaky & 27.19 & 7.49  & 7.04  & 3.88  & 7.04  \\
                   & Faroese-all      & 27.67 & 7.81  & 7.54  & 4.1   & 7.54  \\
\midrule
Afrikaans-nonleaky & Faroese-leaky    & 47.44 & 16.28 & 7.43  & 5.57  & 7.43  \\
                   & Faroese-nonleaky & 39.78 & 15.99 & 6.96  & 5.47  & 6.96  \\
                   & Faroese-all      & 39.97 & 16    & 6.97  & 5.47  & 6.97  \\
\midrule
Afrikaans-all      & Faroese-leaky    & 45.58 & 17.21 & 22.8  & 10.42 & 22.8  \\
                   & Faroese-nonleaky & 40.51 & 16.34 & 10.03 & 5.82  & 10.03 \\
                   & Faroese-all      & 40.64 & 16.36 & 10.37 & 5.94  & 10.37 \\
\midrule
Afrikaans-diverse  & Faroese-leaky    & 38.6  & 16.74 & 13.04 & 9.94  & 13.04 \\
                   & Faroese-nonleaky & 36.18 & 16.34 & 8.13  & 6.48  & 8.13  \\
                   & Faroese-all      & 36.24 & 16.35 & 8.26  & 6.57  & 8.26  \\
\bottomrule
\end{tabular}
\end{table*}
\clearpage
\newpage
\begin{table*}
\begin{tabular}{cc|ccccc}
\toprule
\textbf{Train}              & \textbf{Test}             & \textbf{UAS}   & \textbf{LAS}   & \textbf{CLAS}  & \textbf{MLAS}  & \textbf{BLEX}  \\
\midrule
Danish-leaky       & Faroese-leaky    & 71.33 & 53.85 & 55.73 & 47.83 & 55.73 \\
                   & Faroese-nonleaky & 49.86 & 35.97 & 34.74 & 27.31 & 34.74 \\
                   & Faroese-all      & 50.93 & 36.86 & 35.86 & 28.41 & 35.86 \\
\midrule
Danish-nonleaky    & Faroese-leaky    & 73.43 & 60.37 & 65.05 & 57.37 & 65.05 \\
                   & Faroese-nonleaky & 64.1  & 52.26 & 47.85 & 42.12 & 47.85 \\
                   & Faroese-all      & 64.57 & 52.66 & 48.77 & 42.94 & 48.77 \\
\midrule
Danish-all         & Faroese-leaky    & 69.46 & 59.21 & 64.08 & 54.69 & 64.08 \\
                   & Faroese-nonleaky & 62.52 & 51.63 & 48.34 & 42.12 & 48.34 \\
                   & Faroese-all      & 62.86 & 52    & 49.18 & 42.79 & 49.18 \\
\midrule
Danish-diverse     & Faroese-leaky    & 68.53 & 57.34 & 62.75 & 52.23 & 62.75 \\
                   & Faroese-nonleaky & 62.92 & 51.13 & 48.3  & 41.82 & 48.3  \\
                   & Faroese-all      & 63.2  & 51.43 & 49.08 & 42.38 & 49.08 \\
\midrule
Icelandic-leaky    & Faroese-leaky    & 88.07 & 80.97 & 74.1  & 69.54 & 74.1  \\
                   & Faroese-nonleaky & 72.77 & 65.78 & 53.78 & 50.26 & 53.78 \\
                   & Faroese-all      & 74.02 & 67.02 & 55.54 & 51.93 & 55.54 \\
\midrule
Icelandic-nonleaky & Faroese-leaky    & 93.32 & 89.49 & 84.75 & 83.55 & 84.75 \\
                   & Faroese-nonleaky & 84.18 & 79.21 & 70.54 & 67.5  & 70.54 \\
                   & Faroese-all      & 84.93 & 80.04 & 71.77 & 68.9  & 71.77 \\
\midrule
Icelandic-all      & Faroese-leaky    & 91.05 & 87.64 & 82.93 & 80.53 & 82.93 \\
                   & Faroese-nonleaky & 84.27 & 79.97 & 71.36 & 68.49 & 71.36 \\
                   & Faroese-all      & 84.82 & 80.6  & 72.37 & 69.54 & 72.37 \\
\midrule
Icelandic-diverse  & Faroese-leaky    & 91.76 & 87.36 & 82.07 & 80.39 & 82.07 \\
                   & Faroese-nonleaky & 84.89 & 80.49 & 72.7  & 69.88 & 72.7  \\
                   & Faroese-all      & 85.45 & 81.05 & 73.51 & 70.79 & 73.51 \\
\midrule
Norwegian-leaky    & Faroese-leaky    & 72.46 & 61.41 & 65.53 & 61.8  & 65.53 \\
                   & Faroese-nonleaky & 52.78 & 43.48 & 36.09 & 32.2  & 36.09 \\
                   & Faroese-all      & 54.04 & 44.62 & 38.09 & 34.21 & 38.09 \\
\midrule
Norwegian-nonleaky & Faroese-leaky    & 73.01 & 62.5  & 67.39 & 62.42 & 67.39 \\
                   & Faroese-nonleaky & 59.37 & 49.59 & 47    & 43.19 & 47    \\
                   & Faroese-all      & 60.24 & 50.42 & 48.4  & 44.51 & 48.4  \\
\midrule
Norwegian-all      & Faroese-leaky    & 73.01 & 61.59 & 65.54 & 59.69 & 65.54 \\
                   & Faroese-nonleaky & 61.12 & 49.49 & 45.64 & 42.46 & 45.64 \\
                   & Faroese-all      & 61.88 & 50.27 & 47.01 & 43.64 & 47.01 \\
\midrule
Norwegian-diverse  & Faroese-leaky    & 70.83 & 62.86 & 67.08 & 62.11 & 67.08 \\
                   & Faroese-nonleaky & 61.12 & 51.3  & 48.71 & 45.04 & 48.71 \\
                   & Faroese-all      & 61.74 & 52.04 & 49.96 & 46.21 & 49.96 \\
\bottomrule
\end{tabular}
\end{table*}
\clearpage
\newpage
\begin{table*}
\begin{tabular}{cc|ccccc}
\toprule
\textbf{Train}              & \textbf{Test}             & \textbf{UAS}   & \textbf{LAS}   & \textbf{CLAS}  & \textbf{MLAS}  & \textbf{BLEX}  \\
\midrule
Swedish-leaky      & Faroese-leaky    & 60.58 & 32.75 & 38.81 & 27.29 & 38.81 \\
                   & Faroese-nonleaky & 42.92 & 15.44 & 17.44 & 14.72 & 17.44 \\
                   & Faroese-all      & 43.63 & 16.13 & 18.33 & 15.24 & 18.33 \\
\midrule
Swedish-nonleaky   & Faroese-leaky    & 75.07 & 61.74 & 64.22 & 58.33 & 64.22 \\
                   & Faroese-nonleaky & 57.22 & 41.87 & 38.77 & 33.37 & 38.77 \\
                   & Faroese-all      & 57.94 & 42.67 & 39.83 & 34.41 & 39.83 \\
\midrule
Swedish-all        & Faroese-leaky    & 68.99 & 55.07 & 57.35 & 49.76 & 57.35 \\
                   & Faroese-nonleaky & 49.95 & 34.85 & 31.8  & 26.79 & 31.8  \\
                   & Faroese-all      & 50.71 & 35.65 & 32.86 & 27.74 & 32.86 \\
\midrule
Swedish-diverse    & Faroese-leaky    & 55.36 & 44.06 & 46.04 & 30.69 & 46.04 \\
                   & Faroese-nonleaky & 40.15 & 29.34 & 22.84 & 18.06 & 22.84 \\
                   & Faroese-all      & 40.76 & 29.93 & 23.82 & 18.59 & 23. \\
\bottomrule
\end{tabular}
\caption{Results of our experiment on bilingual transfer for Faroese.}
\label{tab:faroese}
\end{table*}

\end{document}